\documentclass{article}
\usepackage{spconf,amsmath,graphicx}
\usepackage[section]{placeins}
\usepackage{amssymb}
\usepackage{multirow}

\title{TEMPORAL MEMORY ATTENTION FOR VIDEO SEMANTIC SEGMENTATION}
\name{Hao Wang\textsuperscript{\rm{1,2}}, Weining Wang\textsuperscript{\rm{1,2}},  Jing Liu\textsuperscript{\rm{1,2}}}
\address{\textsuperscript{1} National Laboratory of Pattern Recognition, Institute of Automation, Chinese Academy of Sciences \\
	\textsuperscript{2} School of Artificial Intelligence, University of Chinese Academy of Sciences}
\begin{document}
	%
	\maketitle
	\begin{abstract}
		
		
		Video semantic segmentation requires to utilize the complex temporal relations between  frames of the video sequence. Previous works usually exploit accurate optical flow to leverage the temporal relations, which suffer much from heavy computational cost. In this paper, we propose a Temporal Memory Attention Network (TMANet) to adaptively integrate the long-range temporal relations over the video sequence based on the self-attention mechanism without exhaustive optical flow prediction. Specially, we construct a memory using several past frames to store the temporal information of the current frame. We then propose a temporal memory attention module to capture the relation between the current frame and the memory to enhance the representation of the current frame. Our method achieves new state-of-the-art performances on two challenging video semantic segmentation datasets, particularly 80.3\% mIoU on Cityscapes and 76.5\% mIoU on CamVid with ResNet-50.
		
	\end{abstract}
	\section{Introduction}
	\label{sec:intro}
	Image semantic segmentation is a dense prediction task that needs to predict a category label for each pixel of a given image.  Video semantic segmentation is a much more challenging task, which needs to assign a category label for each pixel in each frame of a given video sequence. 
	
	Video semantic segmentation is an important task for visual understanding, which has attracted a lot of attention from the research community \cite{CLK2016, DFF2017, LVS2018, TDNet2019}. The most straightforward solution for video semantic segmentation is to apply an image semantic segmentation model to each frame of the videos as image semantic segmentation does. However, video frames have strong relation with each other. 
	Simply applying an image segmentation model on a video sequence frame by frame doesn't make full use of the temporal relation between video frames. 
	Modeling the temporal relation of video frames  will improve the performance of the video segmentation model. 
	Previous works building the temporal relation of a video sequence can be categorized into two streams: optical-flow-based methods and non-optical-flow-based methods.
	
	\begin{figure}[thb!]
		\begin{center}
			\includegraphics[width=0.95\linewidth]{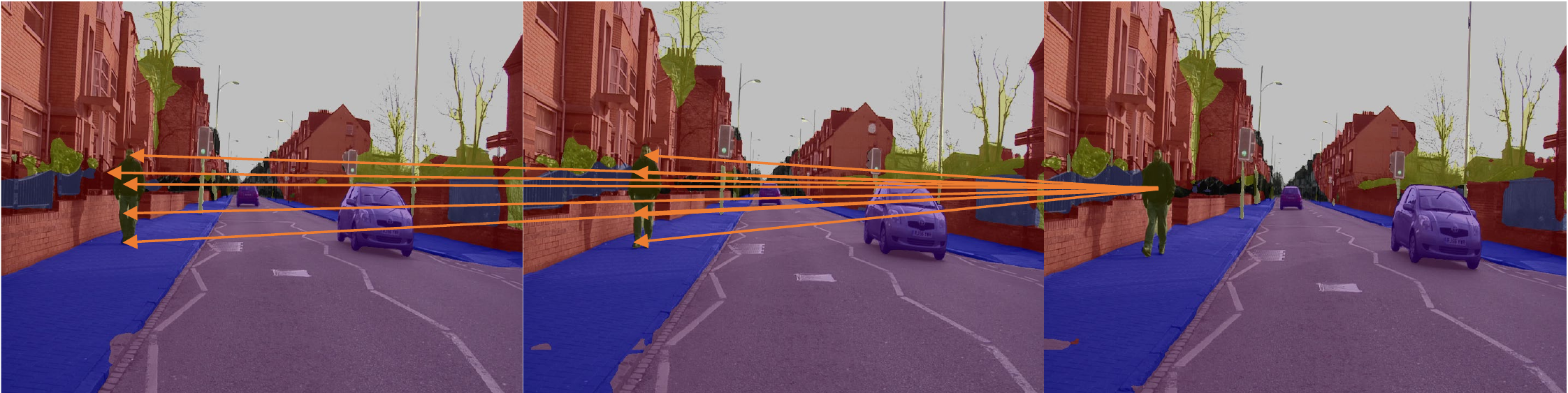}
		\end{center}
		\vspace{-0.5cm}
		\caption{An example of the behavior of TMANet. It collects related information from the previous frames to enhance the representation of the current frame. The orange arrows represent the highly related positions between the frames.}
		\vspace{-0.5cm}
		\label{example}
	\end{figure}
	
	Optical flow represents the motion of an object between consecutive frames. The optical-flow-based methods \cite{CLK2016, DFF2017, GRFP2018, LVS2018} usually contain two networks: 1) an optical flow network, which predicts the motion of objects between consecutive frames by a well pre-trained optical flow network (\emph{e.g.} FlowNet-2.0 \cite{FlowNet2.02017}), and 2) a segmentation network, which generates the segmentation results for the pre-defined key frame and  uses the predicted optical flow to propagate the segmentation result from the key frame to other frames.
	%
	Optical-flow based methods share the same point that video segmentation model needs high-quality optical flow predictions, and poor optical flow predictions will lead to poor segmentation results.
	
	The non-optical-flow-based methods raise a new direction to generate the video representation and achieve better performance recently. 
	Per-frame prediction method \cite{efficient2020} on video semantic segmentation introduces a novel temporal consistency loss to improve the temporal consistency of video prediction and employs a light model with knowledge distillation to retain high performance and attain high inference speed simultaneously. 
	TDNet \cite{TDNet2019} proposes to distribute several sub-networks over sequential frames and recompose the extracted features for segmentation via an attention propagation module.
	The non-optical-flow based methods discard the optical flow prediction, which is more efficient for video semantic segmentation. 
	Our proposed method  belongs to non-optical-flow-based method.
	
	\begin{figure*}[thb!]
		\begin{center}
			\includegraphics[width=0.95\linewidth]{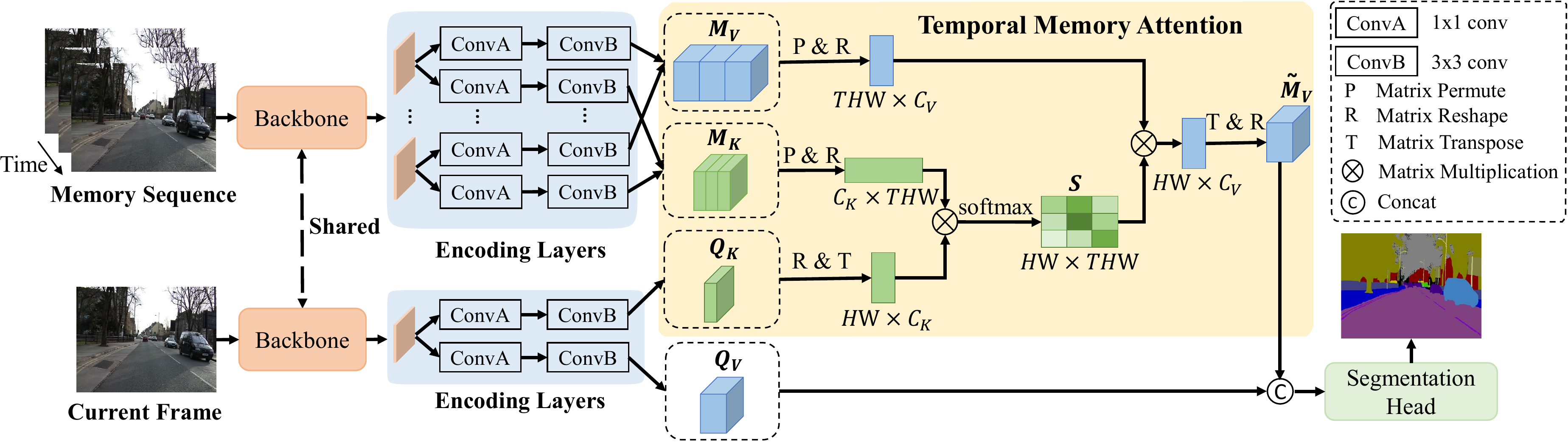}
		\end{center}
		\vspace{-0.5cm}
		\caption{Illustration of our proposed TMANet. We select $T$ frames from a given video as the memory sequence.  The current frame and memory sequence are fed into a shared backbone to extract features. The encoding layers further embed the features to keys and values. The Temporal Memory Attention module captures temporal relation between $Q_K$, $M_K$ and $M_V$,  generating an enhanced memory embedding $\tilde{M}_V$. The embedding of current frame $Q_V$ is concatenated with $\tilde{M}_V$ to generate final segmentation result through a segmentation head. Best viewed in color.}
		\vspace{-0.36cm}
		\label{overview}
	\end{figure*}
	
	Memory networks have been introduced to enhance the reasoning ability of the model in VideoQA \cite{motion2018, heterogeneous2019} and video object segmentation \cite{STA2019, faset_vos2020, enhanced2019}, but have never been introduced in video semantic segmentation as we know. 
	\cite{motion2018}  uses episodic memory to conduct multiple cycles of inference by interacting the question with video features conditioned on current memory. 
	STA \cite{STA2019} designs a spatial-temporal attention mechanism to capture the temporal information for video object segmentation. Memory networks utilize a memory component to store and retrieve information required by the query from the memory. 
	
	In video representation, it is straightforward to construct a memory that consists of the previous frames and a query represents the current frame. 
	Then, we can retrieve information from the previous frames by computing the correlation between the previous frames and the current frame to enhance the representation of the current frame.  
	Motivated by this, we propose a Temporal Memory Attention network (TMANet) to better capture the temporal relation of video frames and enhance the video representation without the help of optical-flow. 
	Take the street scene in Fig.\ref{example} as an example, the person appearing on the current frame also appears in the previous frames, which exists high relationships between adjacent frames. Our model aims to adaptively integrate similar information from the previous frames, thus enhances the representation of the current frame and improves the segmentation results.
	
	Our main contributions are as follows: 
	\textbf{(1)} 
	We propose a novel Temporal Memory Attention Network, which is the first work applying the memory and self-attention mechanism in video semantic segmentation. 
	\textbf{(2)} We design a novel Temporal Memory Attention module to capture the temporal correlation in the video sequence efficiently. 
	\textbf{(3)} The proposed method achieves new state-of-the-art performances on two challenging datasets, namely Cityscapes and CamVid.
	
	\section{Methodology}
	\label{sec:approach}

	\subsection{Overview}

	Given a video sequence that contains multiple frames where one frame is annotated with labels, we consider the previous frames without annotation labels as the memory frames and the current frame with annotation label as query frame. It should be noted that the memory contains multiple frames, while the query contains one frame. Both the memory and the query frame are then fed into a shared backbone to extract features following previous works \cite{TDNet2019, PSPNet2017, DANet2019}. The output of the backbone is of high dimension but in low resolution. To reduce computational cost and encode different representation of the memory and the query, the extracted features from the backbone are fed into encoding layers for channel reduction and feature encoding. The key feature is learned to encode visual semantics for matching robust appearance variations, the value feature stores detailed information for producing semantic prediction, and the number of channel in the key feature is much smaller than that of the value feature. Next, the key and value feature go through our proposed Temporal Memory Attention (TMA) module to build the long-range temporal context information. Then, the value features of query is combined with the long-range temporal context information to enhance the query representation. After feature aggregation, a segmentation head is followed to output the final segmentation result for the current frame.

	As illustrated in Fig.\ref{overview}, given a memory sequence containing $T$ frames and a query with a single frame $X \in R^{3 \times H \times W}$, we concatenate the memory frames along the temporal dimension to get a new memory $M\in R^{T\times3 \times H \times W}$. First, features are extracted via a shared deep backbone. Then, we feed them into different encoding layers to generate features with different semantic information, $M_K\in R^{T \times C_K \times H \times W}$, $M_V \in R^{T \times C_V \times H \times W}$ for memory and $Q_K\in R^{C_K \times H \times W}$ , $Q_V \in R^{C_V \times H \times W}$ for query. After that, the key and value are input to the Temporal Memory Attention module to capture the long-range temporal relations. We add a simple feature aggregation following \cite{DANet2019, STA2019} to aggregate the temporal information in memory and important information in query. Finally, We add a segmentation head implemented by 1x1 convolution to generate segmentation map ($ R^{C \times H \times W}$), where $C$ is the number of classes.
	
	\subsection{Encoding Layer}
	Directly using the original output of the backbone is computationally expensive because of the high-dimensional channel.
	The simplest way for channel reduction is applying a 1x1 convolution on the feature maps. However, 1x1 convolution is not able to capture the spatial information and leads to performance decreasing. 
	The 3x3 convolution or larger kernel can capture spatial information with a larger receptive field, but it will bring more parameters and computational cost. 
	Therefore, we propose to apply a 1x1 convolution for channel reduction and add a 3x3 convolution for spatial information encoding to balance the performance and computation.
	
	\subsection{Temporal Memory Attention Module}
	
	As for images, long-range context refers to the relation between a unique pixel and other pixels \cite{DANet2019}, while the long-range context of videos is the relation between different frames \cite{nonlocal2018, STA2019}. As represented in Fig.\ref{overview}, we propose a Temporal Memory Attention module to build the temporal relations of video frames for video semantic segmentation.
	
	After embedding the memory sequence as mentioned above, we accordingly obtain $T$ key features and $T$ value features. 
	We then concatenate them along the temporal dimension generating a 4-dimension matrix, and then permute and reshape them to $M_K\in R^{C_K \times M}$ and $M_V\in R^{M \times C_V}$, respectively.  
	$M = T \times H \times W$ is the number of pixels in the memory.
	Similarly, we reshape and transpose the key of query to $Q_K\in R^{N \times C_K}$, where $N = H \times W$ is the number of pixels in the query.
	Next, we multiply $M_K$ and $Q_K$, and then apply a softmax layer to calculate the temporal memory attention $S\in R^{N\times M}$,
	\vspace{-0.2cm}
	\begin{equation}
		S_{ij}=\frac{exp(Q_K^i \cdot M_K^j)}{\sum_{j=1}^{M}exp(Q_K^i \cdot M_K^j)} \label{Softmax}
		\vspace{-0.2cm}
	\end{equation}
	where $S_{ij}$ measures the impact of the $i^{th}$ position in the key of query on the $j^{th}$ position in the key of memory.  
	It should be noted that larger impact from the query to the memory indicates greater relation between them.
	After obtaining the temporal attention map $S$, we multiply $S$ and $M_V$ to integrate the temporal relation to memory, thus enhancing the embedding of memory.
	
	\subsection{Feature Aggregation}
	
	After obtaining the long-range temporal context information via temporal memory attention module, 
	we combine the long-range temporal context information with the information from current frame, as follows:
	\vspace{-0.2cm}
	\begin{equation}
		f=\Theta (\tilde{M_V}, Q_V)
		\vspace{-0.2cm}
	\end{equation}
	where \textit{f} is the aggregated feature, and \textit{$\Theta$} is the employed feature aggregation method.
	
	Feature aggregation can be implemented by a decoder structure \cite{deeplabv3+2018} , feature concatenation or feature summation. In this paper, we employ feature concatenation for simplicity. After feature aggregation, we exploit a segmentation head to generate the final segmentation result for the current frame.
	
	\section{EXPERIMENTS}
	\label{sec:experiments}
	\subsection{Dataset and Implementation Details}
	\label{ssec:dataset and implementation}
	To evaluate our proposed method, we carry out comprehensive experiments on two benchmark datasets Cityscapes\cite{Cityscapes} and CamVid\cite{Camvid}.
	
	Cityscapes \cite{Cityscapes} contains 5000 high-quality fine annotated images, which can be split into 2975, 500 and 1275 snippets for training, validation and testing, respectively. Each snippet contains 30 frames, and only the 20$^{th}$ frame of each snippet is annotated with 19 classes for semantic segmentation. CamVid \cite{Camvid} contains 4 videos with 11 category labels for semantic segmentation and is annotated every 30 frames. The annotated frames are grouped into 467, 100 and 233 snippets for training, validation and testing, respectively. We adopt mean Intersection-over-Union (mIoU) as our evaluation metric on Cityscapes and CamVid.
	
	We implement our method based on PyTorch on 4 GPUs of Tesla V100. Inspired by \cite{deeplab2017}, we employ the poly learning rate policy and employ SGD as the optimizer, where the initial learning rate is multiplied by $(1 - \frac{iter}{total\_iter})^{0.9}$ for each iteration. 
	Momentum and weight decay are set to 0.9 and 5e-4  for all experiments on Cityscapes and CamVid. 
	We train our model with Sync-BN \cite{encoding2018}, where batch size and learning rate are set to 8 and 0.01 for both datasets, respectively. 
	We set the total iteration to 80,000 for all experiments. For data augmentation, we apply random resize with a ratio between 0.5 and 2, random cropping (768x768, 640x640 for Cityscapes and CamVid respectively) and random horizontal flipping for input images and sequences for all experiments. We apply sliding window strategy to generate video snippets in the testing stage. Following \cite{PSPNet2017}, we add the auxiliary segmentation loss at the low-level feature of the backbone (\emph{e.g.} the stage 3 output of ResNet). We adopt the above settings for all experiments if without specific clarification.
	
	\subsection{Ablation Study}
	\label{ssec:ablations}
	
	\begin{table}
		\caption{Comparison results of different channel numbers of key features and memory lengths on the Cityscapes validation set. Sequence2 denotes 2 frames in the Memory. Key256 denotes the channel number of key is 256.}
		\vspace{-0.3cm}
		\begin{center}
			\begin{tabular}{l | c }
				\hline
				Method & mIoU (\%) \\
				\hline\hline
				Baseline & 70.69 \\
				\hline
				Sequence2-Key256 & 77.77 \\
				Sequence2-Key128 & 77.95 \\
				Sequence2-Key64 & 77.87 \\
				Sequence2-Key32 & 77.65 \\
				\hline
				Sequence1-Key64 & 78.08 \\ 
				Sequence4-Key64 & 78.26 \\ 
				Sequence6-Key64 & 78.28 \\
				\hline
			\end{tabular}
		\end{center}
		\vspace{-0.5cm}
		\label{ablation 01}
	\end{table}
	
	\begin{table}[t]
		\vspace{-0.3cm}
		\caption{Comparison results of different feature aggregation methods and sampling methods (default is random) on the Cityscapes validation set.}
		\vspace{-0.3cm}
		\begin{center}
			\begin{tabular}{l | c | c}
				\hline
				Method & Sample & mIoU (\%) \\
				\hline\hline
				Sequence4-Key64, concat & random & 78.39 \\
				Sequence4-Key64, concat & continuous & 78.45 \\
				Sequence4-Key64, sum & random & 78.18 \\
				Sequence4-Key64, sum & continuous & 78.33 \\
				\hline
			\end{tabular}
		\end{center}
		\vspace{-0.5cm}
		\label{ablation 03}
	\end{table}
	
	\begin{table}[t]
		\caption{Comparison results of different encoding layers on the Cityscapes validation set.}
		\vspace{-0.3cm}
		\begin{center}
			\begin{tabular}{l | c }
				\hline
				Method & mIoU (\%) \\
				\hline\hline
				Sequence4-Key64, 3x3 conv & 78.26 \\
				Sequence4-Key64, 1x1 conv & 77.88 \\
				Sequence4-Key64, 1x1 conv, 3x3 conv & 78.39 \\
				\hline
			\end{tabular}
		\end{center}
		\vspace{-0.5cm}
		\label{ablation 02}
	\end{table}

	\begin{table}
		\vspace{-0.3cm}
		\caption{Comparison results with state-of-the-arts on Cityscapes and CamVid validation set.}
		\vspace{-0.3cm}
		\begin{center}
    		\begin{tabular}{l | c | c | c}
    			\hline
    			\multirow{2}{*}{Method} & \multicolumn{2}{c|}{mIoU (\%)} & \multirow{2}{*}{GFLOPs} \\
    			& Cityscapes & CamVid & \\
    			\hline\hline
    			DFF \cite{DFF2017} & 69.2 & - & $>$919 \\
    			GRFP \cite{GRFP2018} & 73.6 & 66.1 & - \\
    			Netwarp \cite{netwarping2017} & - & 67.1 & $>$919 \\
    			LVS \cite{LVS2018} & 76.8 & - & - \\
    			\hline
    			TDNet-50 \cite{TDNet2019} & 79.9 & 76.0 & 1082 \\
    			\hline
    			\textbf{Ours-50} & \textbf{80.3} & \textbf{76.5} & \bf{754} \\
    			\hline
    		\end{tabular}
    	\end{center}
		\vspace{-0.7cm}
		\label{sota}
	\end{table}
	
	All the ablation experiments are conducted on the Cityscapes dataset. We use FCN-50\cite{FCN2015} as our baseline. To save computational resources and training time, we adopt ResNet-50 as the backbone and set output stride to 16 for all ablation experiments. 
	
	Following \cite{STA2019}, we set the channel of value features in both memory and query as four times than that of key features (\emph{e.g.} when the channel of key features is set to 64, the channel of value features is set to 256). 
	Besides, it is important to determine how many past frames should be selected into memory. 
	We conduct experiments to analyze different numbers of channel and different memory lengths. As shown in Table \ref{ablation 01}, we can observe that a significant improvement from 77.52 to 78.28 is obtained when the length of memory increases from 2 to 4. While when the length of memory increases to 6, the improvement is too slight to be ignored. 
	It can be interpreted as information redundancy that the memory storing the information of 4 frames is enough for the feature representation enhancement of the current frame.
	Though the model performs best when the channel number is set to 128, we choose 64 as the number of channel in key features for computational efficiency which has similar performance as 128 channels.  
	
	To build the memory, we need to select multiple frames from the past video sequence.
	There exist two selection methods as follows: 1) random selecting multiple frames from the past video sequence (random selecting $n$ frames from last 10 frames), 2) continuously selecting multiple frames from the past video sequence. 
	As shown in Table \ref{ablation 03},  the continuous selecting strategy performs better than random selecting strategy. The possible reason is that random selecting strategy may involve some long-range relation which is harmful to the current frame representation because of the long distance from the current frame. While the continuous selecting strategy selects multiple frames from the current frame continuously, the representation between frames is highly related, which will enhance the feature representation of the current frame. We also analyze different feature aggregation methods, e.g. concatenation and summation. As shown in Table \ref{ablation 03}, feature concatenation performs better than feature summation. The main reason appears to be that concatenated features involve more channels and can represent more information. 
	
The encoding layer plays an important role in the framework, thus we also compare different encoding layers and the results are listed in Table \ref{ablation 02}. It can be seen that a combination of  1x1 convolution and  3x3 convolution performs best than other configurations.

	
	\subsection{Comparison with State-of-the-arts}
	\label{ssec:sota}
Table \ref{sota} shows the performance and GFLOPs of our method and other state-of-the-art methods. Considering the inference time varies from different hardware environments, we provide the computational cost of GFLOPs for fair comparison. Compared with other optical-flow based methods and non optical-flow based methods, our method  achieves better performance on both Cityscapes and Camvid datasets with lower computational cost.

	\section{CONCLUSIONS}
	\label{sec:conclusions}
	
	In this paper, we propose a Temporal Memory Attention Network (TMANet) for video semantic segmentation, which is the first work using memory and self-attention to build the temporal relation in video semantic segmentation. 
	Specially, we introduce a Temporal Memory Attention module to capture the temporal relations between frames. 
	Our method achieves state-of-the-art performance on Cityscapes and CamVid dataset without complicated testing augmented skills. 
	In the future, we will continue to decrease the computation complexity and enhance the efficiency of the model.
	
	\section{ACKNOWLEDGEMENTS}
	\label{sec:conclusions}
	This work was supported by National Natural Science Foundation of China (61922086, 61872366) and Beijing Natural Science Foundation (4192059, JQ20022).
	\bibliographystyle{IEEEbib}
	\bibliography{ICIP}

\begin{thebibliography}{10}

\bibitem{CLK2016}
Evan Shelhamer, Kate Rakelly, Judy Hoffman, and Trevor Darrell,
\newblock ``Clockwork convnets for video semantic segmentation,''
\newblock in {\em ECCV}, 2016, pp. 852--868.

\bibitem{DFF2017}
Xizhou Zhu, Yuwen Xiong, Jifeng Dai, Lu~Yuan, and Yichen Wei,
\newblock ``Deep feature flow for video recognition,''
\newblock in {\em CVPR}, 2017, pp. 2349--2358.

\bibitem{LVS2018}
Yule Li, Jianping Shi, and Dahua Lin,
\newblock ``Low-latency video semantic segmentation,''
\newblock in {\em CVPR}, 2018, pp. 5997--6005.

\bibitem{TDNet2019}
Ping Hu, Fabian Caba, Oliver Wang, Zhe Lin, Stan Sclaroff, and Federico
  Perazzi,
\newblock ``Temporally distributed networks for fast video semantic
  segmentation,''
\newblock in {\em CVPR}, 2020, pp. 8818--8827.

\bibitem{GRFP2018}
David Nilsson and Cristian Sminchisescu,
\newblock ``Semantic video segmentation by gated recurrent flow propagation,''
\newblock in {\em CVPR}, 2018, pp. 6819--6828.

\bibitem{FlowNet2.02017}
Eddy Ilg, Nikolaus Mayer, Tonmoy Saikia, Margret Keuper, Alexey Dosovitskiy,
  and Thomas Brox,
\newblock ``Flownet 2.0: Evolution of optical flow estimation with deep
  networks,''
\newblock in {\em CVPR}, 2017, pp. 2462--2470.

\bibitem{efficient2020}
Yifan Liu, Chunhua Shen, Changqian Yu, and Jingdong Wang,
\newblock ``Efficient semantic video segmentation with per-frame inference,''
\newblock {\em arXiv preprint arXiv:2002.11433}, 2020.

\bibitem{motion2018}
Jiyang Gao, Runzhou Ge, Kan Chen, and Ram Nevatia,
\newblock ``Motion-appearance co-memory networks for video question
  answering,''
\newblock in {\em CVPR}, 2018, pp. 6576--6585.

\bibitem{heterogeneous2019}
Chenyou Fan, Xiaofan Zhang, Shu Zhang, Wensheng Wang, Chi Zhang, and Heng
  Huang,
\newblock ``Heterogeneous memory enhanced multimodal attention model for video
  question answering,''
\newblock in {\em CVPR}, 2019, pp. 1999--2007.

\bibitem{STA2019}
Seoung~Wug Oh, Joon-Young Lee, Ning Xu, and Seon~Joo Kim,
\newblock ``Video object segmentation using space-time memory networks,''
\newblock in {\em ICCV}, 2019, pp. 9226--9235.

\bibitem{faset_vos2020}
Yu~Li, Zhuoran Shen, and Ying Shan,
\newblock ``Fast video object segmentation using the global context module,''
\newblock {\em arXiv preprint arXiv:2001.11243}, 2020.

\bibitem{enhanced2019}
Zhishan Zhou, Lejian Ren, Pengfei Xiong, Yifei Ji, Peisen Wang, Haoqiang Fan,
  and Si~Liu,
\newblock ``Enhanced memory network for video segmentation,''
\newblock in {\em ICCV}, 2019, pp. 0--0.

\bibitem{PSPNet2017}
Hengshuang Zhao, Jianping Shi, Xiaojuan Qi, Xiaogang Wang, and Jiaya Jia,
\newblock ``Pyramid scene parsing network,''
\newblock in {\em CVPR}, 2017, pp. 2881--2890.

\bibitem{DANet2019}
Jun Fu, Jing Liu, Haijie Tian, Yong Li, Yongjun Bao, Zhiwei Fang, and Hanqing
  Lu,
\newblock ``Dual attention network for scene segmentation,''
\newblock in {\em CVPR}, 2019, pp. 3146--3154.

\bibitem{nonlocal2018}
Xiaolong Wang, Ross Girshick, Abhinav Gupta, and Kaiming He,
\newblock ``Non-local neural networks,''
\newblock in {\em CVPR}, 2018, pp. 7794--7803.

\bibitem{deeplabv3+2018}
Liang-Chieh Chen, Yukun Zhu, George Papandreou, Florian Schroff, and Hartwig
  Adam,
\newblock ``Encoder-decoder with atrous separable convolution for semantic
  image segmentation,''
\newblock in {\em ECCV}, 2018, pp. 801--818.

\bibitem{Cityscapes}
Marius Cordts, Mohamed Omran, Sebastian Ramos, Timo Rehfeld, Markus Enzweiler,
  Rodrigo Benenson, Uwe Franke, Stefan Roth, and Bernt Schiele,
\newblock ``The cityscapes dataset for semantic urban scene understanding,''
\newblock in {\em CVPR}, 2016, pp. 3213--3223.

\bibitem{Camvid}
Gabriel~J Brostow, Jamie Shotton, Julien Fauqueur, and Roberto Cipolla,
\newblock ``Segmentation and recognition using structure from motion point
  clouds,''
\newblock in {\em ECCV}, 2008, pp. 44--57.

\bibitem{deeplab2017}
Liang-Chieh Chen, George Papandreou, Iasonas Kokkinos, Kevin Murphy, and Alan~L
  Yuille,
\newblock ``Deeplab: Semantic image segmentation with deep convolutional nets,
  atrous convolution, and fully connected crfs,''
\newblock {\em TPAMI}, vol. 40, no. 4, pp. 834--848, 2017.

\bibitem{encoding2018}
Hang Zhang, Kristin Dana, Jianping Shi, Zhongyue Zhang, Xiaogang Wang, Ambrish
  Tyagi, and Amit Agrawal,
\newblock ``Context encoding for semantic segmentation,''
\newblock in {\em CVPR}, 2018, pp. 7151--7160.

\bibitem{netwarping2017}
Raghudeep Gadde, Varun Jampani, and Peter~V Gehler,
\newblock ``Semantic video cnns through representation warping,''
\newblock in {\em ICCV}, 2017, pp. 4453--4462.

\bibitem{FCN2015}
Jonathan Long, Evan Shelhamer, and Trevor Darrell,
\newblock ``Fully convolutional networks for semantic segmentation,''
\newblock in {\em CVPR}, 2015, pp. 3431--3440.

\end{thebibliography}
	
\end{document}